\documentclass[conference,10pt]{IEEEtran}
\IEEEoverridecommandlockouts
\usepackage{cite}
\usepackage{amsmath,amssymb,amsfonts}
\usepackage{algorithm}
\usepackage{algorithmic}
\usepackage{graphicx}
\usepackage{textcomp}
\usepackage{xcolor}

\usepackage{balance}

\usepackage{float}
\usepackage{breqn}

\makeatletter
\newcommand*{\rom}[1]{\expandafter\@slowromancap\romannumeral #1@}
\makeatother
\def\BibTeX{{\rm B\kern-.05em{\sc i\kern-.025em b}\kern-.08em
    T\kern-.1667em\lower.7ex\hbox{E}\kern-.125emX}}
    
\begin{document}

\sloppy

\title{\Huge \textbf{Hybrid Ant Swarm-Based Data Clustering}
}
% \author{Md Ali Azam, Shawon Dey, Hans D. Mittelmann, Shankarachary Ragi}

\makeatletter
\newcommand{\linebreakand}{%
  \end{@IEEEauthorhalign}
  \hfill\mbox{}\par
  \mbox{}\hfill\begin{@IEEEauthorhalign}
}
\makeatother
% \author{\IEEEauthorblockN{\small Md Ali Azam}
% \IEEEauthorblockA{\small \textit{Electrical Engineering} \\
%     \textit{\small South Dakota School of Mines and Technology}\\
%     azam.ete.ruet@gmail.com}
% \and
% \IEEEauthorblockN{\small Abir Hossen}
% \IEEEauthorblockA{\small \textit{Electrical Engineering} \\
%     \textit{\small South Dakota School of Mines and Technology}\\
%     abir.hossen@mines.sdsmt.edu}
% \and
% \IEEEauthorblockN{\small Md hafizur Rahman}
% \IEEEauthorblockA{\small \textit{Electrical Engineering} \\
%     \textit{\small South Dakota School of Mines and Technology}\\
%     mdhafizur.rahman@mines.sdsmt.edu}
% }
\author{
\IEEEauthorblockN{Md Ali Azam,
Md Abir Hossen, Md Hafizur Rahman}
\IEEEauthorblockA{Electrical Engineering\\
South Dakota School of Mines and Technology, Rapid City, SD 57701\\
Email: azam.ete.ruet@gmail.com, abir.hossen@mines.sdsmt.edu, mdhafizur.rahman@mines.sdsmt.edu}
}

\maketitle
\pagestyle{empty}

%\begin{abstract}
%We compare the performance of average consensus and decentralized sensor fusion algorithms for target tracking in different networked sensor scenarios. Specifically, in this paper we show how the performance of these algorithms varies with number of time steps and different sensor network configurations such as the network with number of degrees, edge probability, and total number of edges.
%\end{abstract}
\begin{abstract}
Biologically inspired computing techniques are very effective and useful in many areas of research including data clustering. Ant clustering algorithm is a nature-inspired clustering technique which is extensively studied for over two decades. In this study, we extend the ant clustering algorithm (ACA) to a hybrid ant clustering algorithm (hACA). Specifically, we include a genetic algorithm in standard ACA to extend the hybrid algorithm for better performance. We also introduced novel pick up and drop off rules to speed up the clustering performance. We study the performance of the hACA algorithm and compare with standard ACA as a benchmark.
\end{abstract}


\section{Introduction}
Swarm intelligence has many application such as optimization techniques, robotics, data mining, and communication \cite{blum2008swarm}. Some swarm intelligence techniques e.g., evolutionary algorithms, particle swarm optimization \cite{kennedy1995particle}, genetic algorithms \cite{wang2003genetic}, have revolutionized the field of computing science which are widely used in robotics and data mining. For example, ant swarm-based technique ant colony optimization (ACO) solved complex shortest path problems effectively that made robot navigation easy and simple \cite{dorigo2006ant, ghoseiri2010ant, kumar2018hybridized}. Ant clustering algorithm (ACA) is very useful in data clustering and sorting \cite{Boryczka2008}. 

% big data challenges. motivation
Industries, researchers, medical science, advertising, and governments face a common difficulty with big data analysis \cite{shirkhorshidi2014big}, especially big data clustering. As data collection is a continuous process, there are no alternatives to dealing with big data. Clustering in big data is a very challenging task due to the large volume of data in the dataset. Clustering often becomes more difficult in big data due to heterogeneity and high dimensionality of objects in the dataset. Many data clustering methods have been developed to tackle these challenges such as K-means clustering, mean-shift clustering, agglomerative clustering, and so forth. 

% possible solutions, some literature reviews
Clustering methods can be classified into a few categories, described in \cite{jiang2011new}. Some popular methods of clustering are partitioning method, hierarchical method, density-based method, etc. In a given dataset of $n$ objects, the partitioning method constructs $k$ number of clusters, where $k \leq n$. The partitioning method satisfy two conditions: a) each object must belong to exactly one cluster, b) each cluster contains at least one object. In order to improve the partitioning, the technique requires relocation of objects from one cluster to the other for each iteration which is computationally expensive. 
The hierarchical method deals with decomposition of objects in the dataset. There are two commonly known techniques based on the hierarchical method: agglomerative approach and divisive approach. The agglomerative approach is the most commonly known hierarchical data clustering technique. This is a bottom-up approach where observations start in their own clusters and clusters merge to move up the hierarchy. The divisive approach is a reverse approach to the agglomerative approach. Clustering starts with a single one and splits into multiple ones until a stopping criterion is applied. The hierarchical approaches are effective but computationally expensive, especially for large datasets. 

The partitioning and hierarchical methods are not effective in clusters of arbitrary shapes. The density-based method is efficient in clusters of arbitrary shapes and also robust to noise and outliers. As the name suggests, density-based approach is based on the concept of density. The most used density-based approach DBSCAN approach plays a vital role in finding nonlinear shapes structure. Despite robustness to noise and outliers, density-based approaches are not efficient in high dimensional data and large differences in densities. 

% why ACA better
Biologically inspired swarm intelligence techniques are very effective in clustering. Ant clustering algorithm (ACA) is studied in the past \cite{Labroche2003, Jiang2010, Chen2004} for data mining and clustering applications. The basic idea of ACA is taken from the mechanism of ants to form piles of corpses - cemeteries and larvae. The authors of \cite{Deneubourg1991, Franks1992} did extensive studies of ant-based sorting and clustering behavior primarily based on brood, larvae, and corpses sorting and clustering making a pile of them at a different location. The authors of \cite{Deneubourg1991} represented the behavior of corpses clustering and larvae sorting in ants with algorithms. They introduced general rules of pick up and drop off of objects by an ant. The idea of ant movement in a two-dimensional grid is explained elaborately in \cite{de2007fundamentals}. The ants move one step in any direction (among four possible directions) randomly. Ants pick up an isolated item with a probability and drop at some other location where more of the same type of item is found with some other probability. 

The above-mentioned algorithms might be slow if the nest is large. Their algorithm converges slowly in the clustering of objects if the search space is large. In order to tackle this slow convergence, ACA has been extended in many different ways \cite{Lu2013, Sumangala2020, Chen2020}. The literature lacks extension of ACA to measure the clustering performance, e.g. convergence rate analysis of the ACA. This work extends the ACA where the movement of ants is determined by a genetic algorithm and evaluates the clustering performance of our hybrid ACA.   

% GA ACA advantages

\subsection{Key Contributions}
\begin{itemize}
\item We extend ant clustering algorithm (ACA) to a hybrid ACA
\item We introduce novel pick up and drop off rules for the artificial ants
\item We also compare the performance of the hybrid ACA with ACA as a benchmark
\end{itemize}

The remaining parts of this paper are organized as follows. Section~\ref{sec:problem_spec} provides the hybrid ant clustering algorithm (hACA). Section~\ref{sec:problem_form} includes problem formulation and present the hACA algorithm. Simulation results and performance evaluation is presented in Section~\ref{sec:sim_result}. The paper is concluded with Section~\ref{sec:conc}.

\section{Problem Specification}\label{sec:problem_spec}
The hybrid ant clustering algorithm (hACA) is an extension of the standard ant clustering algorithm (ACA) in terms of ant movement principle in the world and pick up and drop off rules. The central idea of this work is not to demonstrate how actual ants move in the real world. Our objective is to extend ACA in order to speed up the data clustering process. 

\begin{figure}[H]
\centering{\includegraphics[width= 0.6\columnwidth]{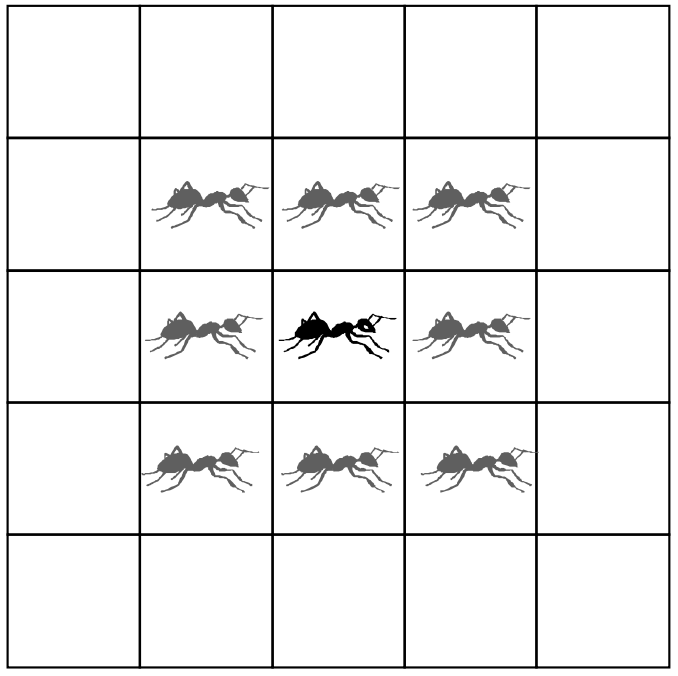}}
\caption{A bi-dimensional grid with $5 \times 5$ cells}
\label{fig: ant_world}
\end{figure}

In the standard ACA, ants are assumed to be moving in a bi-dimensional grid (discrete real-world space) of $m \times m$ cells. The bi-dimensional world of an ant model is proposed by \cite{Lumer1994} where an ant is allowed to travel from one side of the world to the another. The ant can move from its current cell in the grid world to one of its neighboring cells depicted in Figure~\ref{fig: ant_world}. The center cell with the dark ant in the $5 \times 5$ grid shown in the Figure~\ref{fig: ant_world} represents the current location of the ant in the grid world and the grey ants around the center cell represent possible next locations in the grid world. The hACA considers the following extensions to standard ACA.

\textbf{Clustering rules:} The pick up and drop off probabilities of objects in the grid world are also re-defined in the hACA algorithm compared to the standard ACA algorithm proposed in \cite{Deneubourg1991, Lumer1994} which will be discussed more in Section~\ref{sec:problem_form}.

\textbf{Movement of an ant:} The movement of an ant is determined by a genetic algorithm where an ant is allowed to move to any other cell in the grid. This behavior of an artificial ant allows faster data clustering compared to the standard ACA. 

% Ant clustering algorithm (ACA) is a well-known ant-based stochastic clustering algorithm. ACA is widely used to cluster data especially in the field of data analysis. In this problem, I formulate and simulate ACA to determine how ACA performs. Later in this section, I extend ACA to include Genetic Algorithm (mutation to ants’ location) to determine if performance appears to be better than typical ACA. 

% In the first part of this problem, I create a grid of 200x200 cells and place 200 objects randomly in which 100 objects are red and 100 objects are blue. I place 500 ants into the grid randomly such that ants’ choose different location than objects. So, initially ants’ are unloaded. Then I move ants’ randomly to adjacent cells (8 possible adjacent cells). Ants’ pick up an item if corresponding ant is unloaded and cell is occupied with an object. In second iteration, after ants’ moves to random adjacent cell, an ant drop-off the object if the ant is loaded, same object found in any adjacent cells, and corresponding cell is empty. All ants’ follow the same rule, and the process continues number of iterations. Goal is to cluster same colored object together.

% In the second part of the problem, I extend ACA and include GA (mutation to ants’ current location). The reason behind this is, if ants’ are moving only one cell (any neighboring cell) each iteration, chance is very low ants’ can travel long distance that might lead the ACA algorithm to make number of clusters of same object.

\section{Problem Formulation}\label{sec:problem_form}
In this study, an artificial ant is considered to be capable of picking up and dropping off an object if condition satisfies. The general idea is that objects should be picked up and dropped off at some other cells where more of the same kind of objects are present around. Assuming there is only a single type of object present in the environment, an artificial ant will pick up an object with a probability $P_p$ if the ant is unloaded. The probability $P_p$ is given by the following equation. 

\begin{equation}
    P_p = \left(\frac{k_1}{k_1 + f}\right)^2
\end{equation}
where $f$ is the \emph{perceived fraction of items} in the neighborhood of the ant, and $k_1$ is a threshold constant. If $k_1$ is significantly larger than $f$, pick up probability $P_p \approx 1$ which suggests that pick up probability of the object is high if there are not too many objects in the neighborhood. 

The drop off probability $P_d$ is given by the following equation applicable if the ant is already loaded. 

\begin{equation}
    P_d = \left(\frac{f}{k_2 + f}\right)^2
\end{equation}
where $k_2$ is a threshold constant. In this case, if $k_2$ is significantly smaller than $f$, $P_d \approx 1$ which suggests that drop off probability in a dense cluster is high. 

The grid world dimension is considered in a bi-dimensional grid in $\mathbb{R}^2$. $N$ number of artificial ants are placed in the grid $Y \times Z$ randomly to cluster $L$ different types of objects placed randomly initially (total number of ants and objects combined must be less than $YZ$). An ant perceives a surrounding of its current cell $c$ given by $M = s \times s$ neighboring cells. $\mathcal{N}(j)$ is the set of neighboring cells of $i$th artificial ant location $x_i$. $\mathcal{M}(j)$ represents a binary set of neighboring cells of the ant location $x_i$ where $\mathcal{M}(j) = 1$ says an object is present in the $j$th neighbor of the ant. 

This study considers the following equation for \emph{perceived fraction of items} $f$ to calculate the local density in the neighborhood of an ant in the grid world for $i$th artificial ant sitting at current cell $c$. 

\begin{equation}\label{eq3}
    f(x_i) = \sum_{n=1}^{s^2 - 1} \mathcal{N}(j) \mathcal{M}(j)
\end{equation}

\subsection{Pick and Drop Probabilities}
The standard pick up and drop off probabilities are described earlier. We use Equation~\ref{eq3} to calculate the local density in the neighborhood of $i$th ant at cell $c$ which influences the picking and dropping probabilities as follows. 

\begin{equation}
    P_p(x_i) = \left(\frac{k_1}{k_1 + f(x_i)}\right)^2
\end{equation}

\begin{equation}
    P_d(x_i) = \left(\frac{f(x_i)}{k_2 + f(x_i)}\right)^2
\end{equation}

\subsection{The Hybrid Ant Clustering Algorithm (hACA)}
The hybrid ant clustering algorithm (hACA) is described in Algorithm~1. As we described in Section~\ref{sec:problem_spec}, the movement of ants is determined by a genetic algorithm rather than random walk to one of the neighbors. The current cell of an ant in the grid world is represented by a location which includes two variables $(a,b)$ where $a$ and $b$ are row and column respectively. For example, location of the black ant in the Figure~\ref{fig: ant_world} is $(3,3)$. We represent the location of the ant in binary digits. The length of the binary digits is determined by the maximum value of the grid dimension. We perform mutation and recombination to the binary location of an ant to come up with the next location. This movement behavior of an artificial ant allows faster clustering. 

\begin{figure}[H]
\centering{\includegraphics[width= \columnwidth, trim = 10 485 20 70,clip]{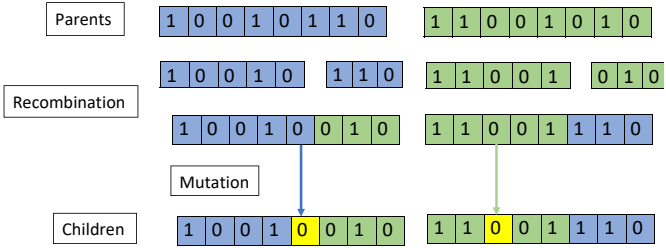}}
\caption{Genetic algorithm applied in hACA}
\label{fig: ga}
\end{figure}

% describe mutation and recombination with image
Location of an ant at the current cell is defined by row and column $(a,b)$ discussed above. Figure~\ref{fig: ga} represents how recombination and mutation are performed on the parents $(a,b)$ to come up with the next location (children) of the ant.

\begin{algorithm}\label{alg}
\caption{Hybrid ant clustering algorithm (hACA)}
\begin{algorithmic} 
\STATE place every ant and all objects in the grid cells randomly
% \STATE place  in the grid cells randomly
\STATE $t \gets 0$
\WHILE{$t \leq \text{MaxIter}$}
\FOR{$i = 1 \text{ to } N$}  
\STATE compute $f(x_i)$
\IF{\text{unloaded ant AND cell occupied by object} $x_i$}
\STATE compute $P_p(x_i)$
\STATE pick up object $x_i$ with probability $P_p(x_i)$
\ELSIF{ant carrying item $x_i$ AND cell empty}
\STATE compute $P_d(x_i)$
\STATE drop object $x_i$ with probability $P_d(x_i)$
\ENDIF
\STATE perform mutation and recombination to ants location 
\ENDFOR
\STATE $t \gets t + 1$
\ENDWHILE
\STATE \textbf{print} location of objects
\end{algorithmic}
\end{algorithm}

\section{Simulation Results}\label{sec:sim_result}
% Assuming that ants move in a bi-dimensional grid, the standard ACA can be
% summarized as in Algorithm 5.3. Let ‘unloaded ant’ be an ant not carrying an
% item, assume f is problem dependent, pp and pd are the probabilities of picking
% up and dropping off an item, respectively, and N is the number of ants. In this
% standard algorithm a cell can only be occupied by a single item.

% describe iteration number

This section provides simulation results for both ACA and hACA algorithms. We create a grid world of $128 \times 128$ cells. We initially place $500$ ants in the grid world represented by black stars. We also place two different types of objects e.g., $100$ red and $100$ blue represented by red and blue circles respectively. Figure~\ref{fig: init} illustrates an initial grid world with ants and objects. The neighboring cells of the $i$th ant are the set of all cells in the $3 \times 3$ grid centering the current cell of the ant. The ants are assumed to be memoryless and do not have access to information e.g., location of other ants in the grid world other than neighboring cells. 

\begin{figure}[H]
\centering{\includegraphics[width= 0.8\columnwidth, trim = 150 260 150 260,clip]{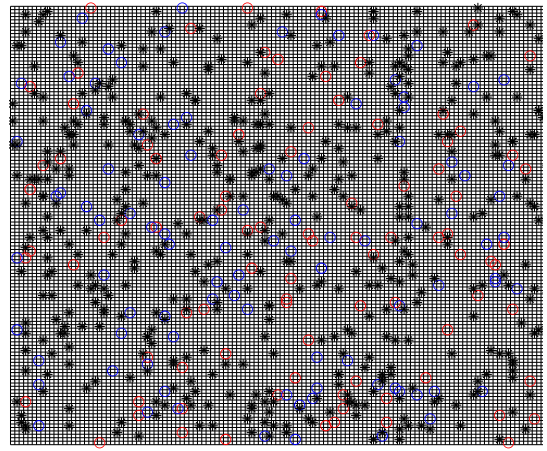}}
\caption{Initial placement of ants and objects in the grid world}
\label{fig: init}
\end{figure}

\begin{figure}[H]
\centering{\includegraphics[width= 0.8\columnwidth, trim = 150 260 150 260,clip]{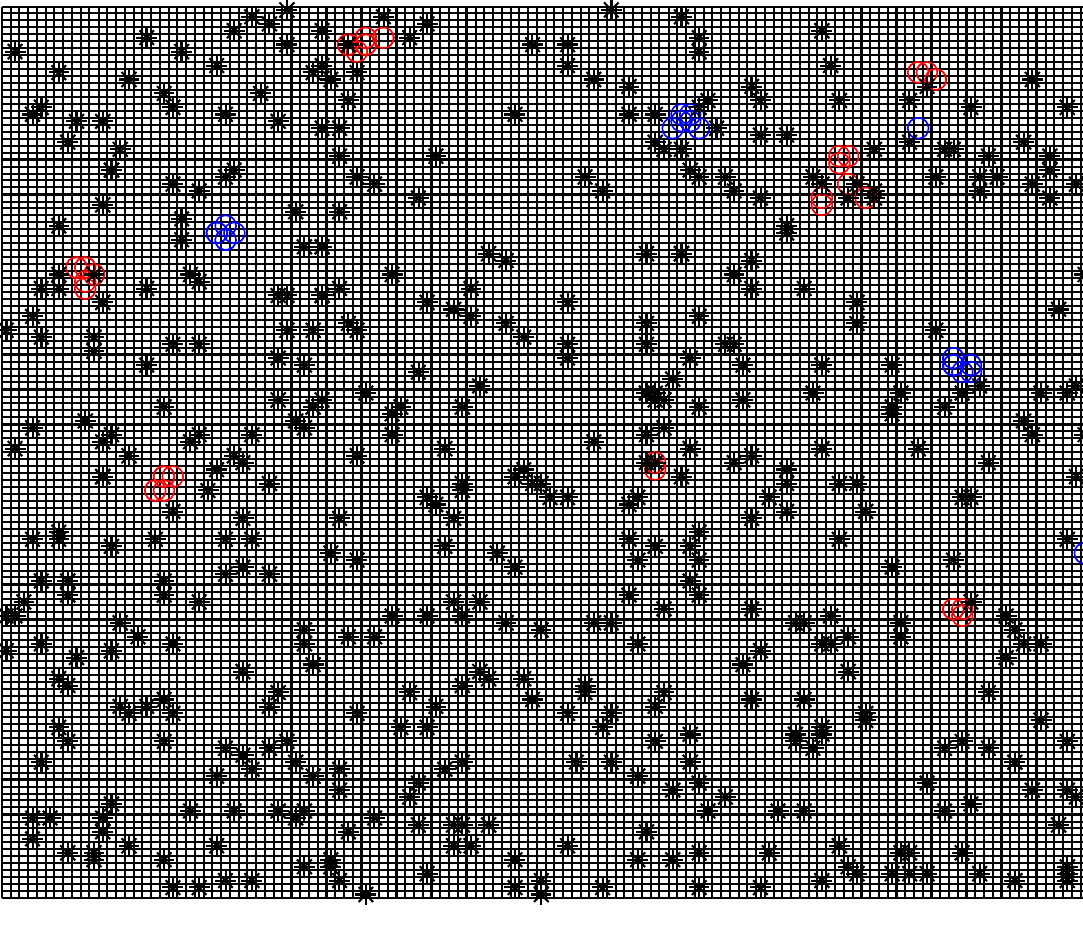}}
\caption{Objects clustered with ACA after $1000$ iterations}
\label{fig: ACA 1000}
\end{figure}

\begin{figure}
\centering{\includegraphics[width= 0.8\columnwidth, trim = 150 260 150 260,clip]{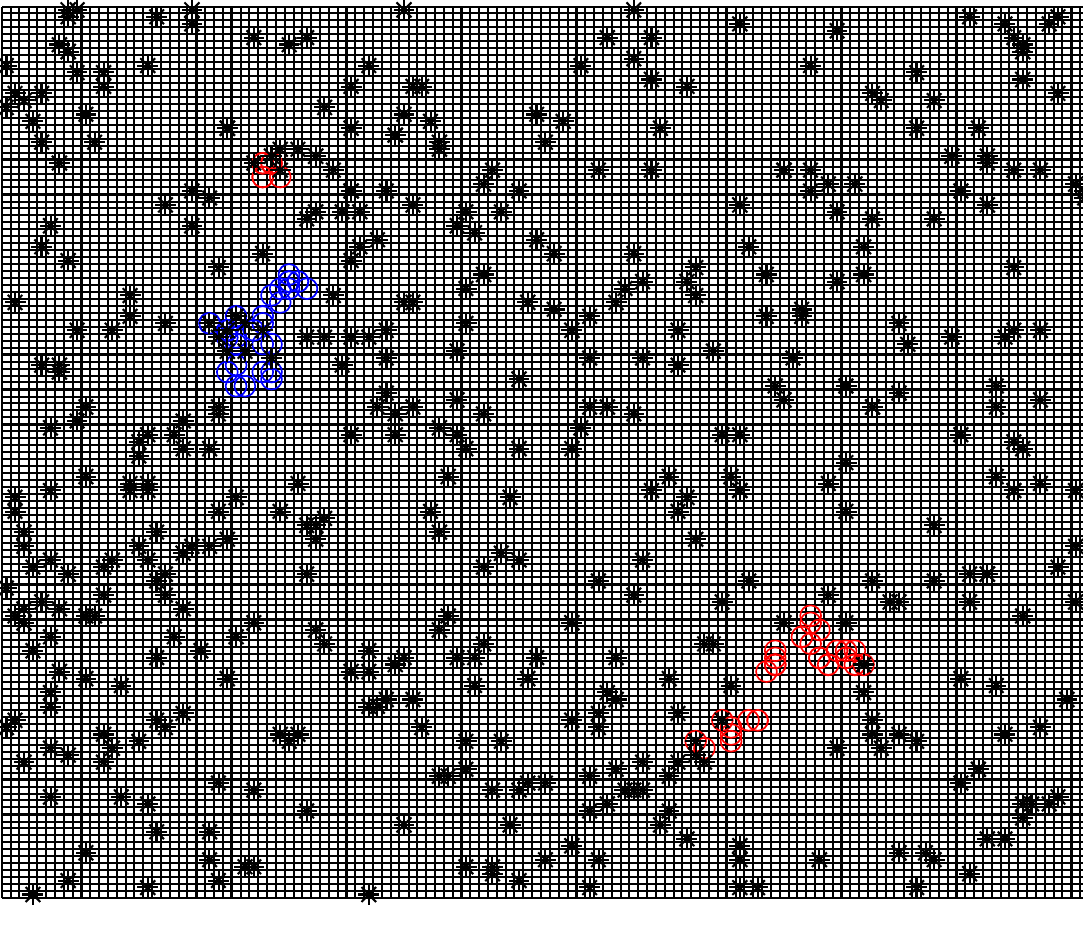}}
\caption{Objects clustered with hACA after $1000$ iterations}
\label{fig: hACA 1000}
\end{figure}

\begin{figure}
\centering{\includegraphics[width= 0.88\columnwidth, trim = 110 235 110 235,clip]{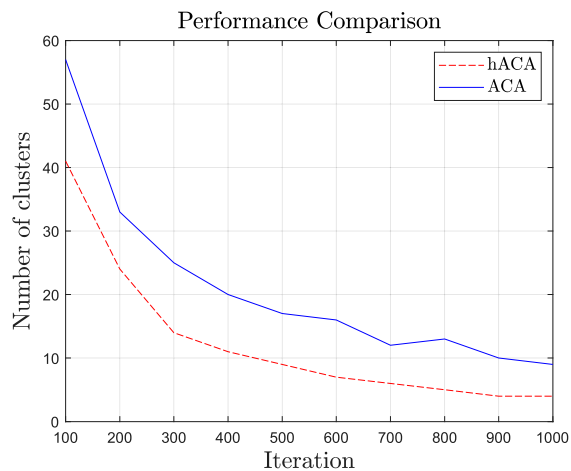}}
\caption{Performance comparison between hACA and standard ACA}
\label{fig: performance comp}
\end{figure}

We apply both ACA and hACA algorithms in this setting for $1000$ iterations. As an ant is considered to be able to move one step (current cell to one of the neighboring cells) for standard ACA, an ant actually cannot move far from its initial location. However, in the case of hACA, ants can take a big step and roam around the entire grid world which allows better clustering performance. Clustering performance is measured in terms of the total number of clusters in the grid after a certain iteration. Figure~\ref{fig: ACA 1000} represents the clustering of both red and blue objects for standard ACA after $1000$ iterations. It is noticeable that ants make a lot of local clusters in this case. However, in the case of hACA algorithm, ants make significantly small number of clusters with big cluster size showed in Figure~\ref{fig: hACA 1000}. We anticipate that in the case of a huge (having millions of grid cells) grid world, hACA will perform further better than the standard ACA algorithm. 

We compare the performance of hACA with the standard ACA algorithm. Figure~\ref{fig: performance comp} illustrates hACA algorithm makes lower number of clusters compared to the standard ACA algorithm for every iteration numbers from $100$ to $1000$.

% memoryless ants
% maybe a study on memoryless vs short term memory

% study on number of iteration vs performance
% define performance 

\section{Conclusions}\label{sec:conc}
In this paper, we extended the standard ant clustering algorithm (ACA) to a hybrid ant clustering algorithm (hACA). We introduced novel object pick up and drop off rules for the artificial ants. We evaluated both the algorithms for $1000$ iterations and found hACA makes small number of clusters compared to standard ACA in terms of iteration number. 

% unlabelled data

\bibliographystyle{IEEEtran}
\bibliography{reference.bib}

\end{document}